\documentclass[letterpaper]{article} 
\usepackage{aaai2026}  
\usepackage{times}  
\usepackage{helvet}  
\usepackage{courier}  
\usepackage[hyphens]{url}  
\usepackage{graphicx} 
\usepackage{natbib}  
\usepackage{caption} 
\usepackage{algorithm}
\usepackage{algorithmic}

\usepackage{newfloat}
\usepackage{listings}
\DeclareCaptionStyle{ruled}{labelfont=normalfont,labelsep=colon,strut=off} 
\floatstyle{ruled}
\newfloat{listing}{tb}{lst}{}
\floatname{listing}{Listing}
\usepackage{booktabs}
\usepackage{tabularx}
\usepackage{multirow}
\usepackage{pifont}
\usepackage[most]{tcolorbox}
\usepackage{arydshln}
\usepackage{booktabs}
\usepackage{siunitx}

\usepackage{tikz}
\usetikzlibrary{arrows.meta, graphs}
\usetikzlibrary{tikzmark, calc}
\usetikzlibrary{decorations.pathreplacing,intersections}

\usepackage{array}
\newcolumntype{L}[1]{>{\raggedright\let\newline\\\arraybackslash\hspace{0pt}}m{#1}}
\newcolumntype{C}[1]{>{\centering\let\newline\\\arraybackslash\hspace{0pt}}m{#1}}
\newcolumntype{R}[1]{>{\raggedleft\let\newline\\\arraybackslash\hspace{0pt}}m{#1}}

\title{GSAP-ERE: Fine-Grained Scholarly Entity and Relation Extraction \\Focused on Machine Learning}
\author{
    Wolfgang Otto\textsuperscript{\rm 1}, 
    Lu Gan\textsuperscript{\rm 1},
    Sharmila Upadhyaya\textsuperscript{\rm 1},
    Saurav Karmakar\textsuperscript{\rm 2},
    Stefan Dietze\textsuperscript{\rm 1,3}
}
\affiliations{
    \textsuperscript{\rm 1}GESIS -- Leibniz Institute for the Social Sciences, Cologne, Germany\\
    \textsuperscript{\rm 2} Individual Researcher \\
    \textsuperscript{\rm 3} Heinrich-Heine-University D\"usseldorf, Germany\\
    wolfgang.otto@gesis.org
}

\begin{document}

\newcommand{\gere}{GSAP-ERE}
\maketitle

\begin{abstract}
Research in Machine Learning (ML) and AI evolves rapidly.
Information Extraction (IE) from scientific publications enables to identify information about research concepts and resources on a large scale and therefore is a pathway to improve understanding and reproducibility of ML-related research.
To extract and connect fine-grained information in ML-related research, e.g. method training and data usage, we introduce \gere.
It is a manually curated fine-grained dataset with 10 entity types and 18 semantically categorized relation types, containing mentions of 63K entities and 35K relations from the full text of 100 ML publications.
We show that our dataset enables fine-tuned models to automatically extract information relevant for downstream tasks ranging from knowledge graph (KG) construction, to monitoring the computational reproducibility of AI research at scale.
Additionally, we use our dataset as a test suite to explore prompting strategies for IE using Large Language Models~(LLM).
We observe that the performance of state-of-the-art LLM prompting methods is largely outperformed by our best fine-tuned baseline model (NER: 80.6\%, RE: 54.0\% for the fine-tuned model vs. NER: 44.4\%, RE: 10.1\% for the LLM). This disparity of performance between supervised models and unsupervised usage of LLMs suggests datasets like \gere~are needed to advance research in the domain of scholarly information extraction.
\end{abstract}

\begin{links}
    \link{Code}{https://data.gesis.org/gsap/gsap-ere#code}
    \link{Dataset}{https://data.gesis.org/gsap/gsap-ere#dataset}
\end{links}


\section{Introduction}
\label{sec:intro}
\noindent Recent studies such as \citeauthor{10.5555/3454287.3454779}~\shortcite{10.5555/3454287.3454779} have documented the declining reproducibility of ML-related research. In addition, determining what models and datasets are state-of-the-art for specific tasks is a very effort-intensive and increasingly challenging process. These problems are elevated not only by poor benchmarking practices \cite{10.1145/3298689.3347058} but also by the surging amount of ML-related research, where understanding the authority, adoption and quality of models and datasets and their interdependencies becomes increasingly hard. Different to other disciplines, prior studies in computer science underline that reproducibility problems are to a large extent caused by poor reporting practices in scholarly publications, where only 4\% of the assessed ML-related publications could be reproduced when the original authors did not respond to clarification requests \cite{JMLR:v22:20-303}.

Scholarly information extraction facilitates mining and understanding of vast amounts of scholarly literature~\cite{Luan2018SciERC,luan-etal-2019-general,jain2020scirex,huitong2023dmdd,zhang2024scier} and usually involves Named Entity Recognition (NER) and Relation Extraction (RE), enabling the identification of entities and their relationships. The joined Entity and Relation Extraction (ERE) enables tasks, such as academic question answering~\cite{dasigi-etal-2021-dataset} and scholarly knowledge graph construction~\cite{viswanathan-etal-2021-citationie}, and thus facilitates reproducibility and reusability of research.

In particular among scholarly entities relevant to AI and ML-related research, e.g. dataset, ML model and task, various dependencies are of great interest to enhance research reproducibility and reusability~\cite{arvan-etal-2022-reproducibility-computational,wonsil2023rep}. 
\emph{Data-model dependencies} describe the datasets used for training and evaluating models; \emph{model-task dependencies} indicate the downstream tasks applied to the trained or fine-tuned models; \emph{inner-entity-type dependencies} capture comparability between scholarly resources or research concepts; and \emph{entity-tracing dependencies} reflect the provenance or source of research artifacts.

However, many existing IE methods build upon datasets with a coarse-grained entity label set~\cite{Luan2018SciERC,zhang2024scier}, leading to an impossible identification of necessary research artifact metadata. Second, despite the huge progress in recent years, LLMs~\cite{qwen2025qwen25technicalreport,touvron2023llamaopenefficientfoundation} still lag behind largely on fine-grained domain-specific IE tasks compared to fine-tuned models~\cite{zhang2024scier}. Hence, it is not suitable to directly apply LLMs for a high-quality fine-grained scholarly IE.
Third, despite of initial effort to a fine-grained scholarly NER dataset like GSAP-NER~\cite{otto-etal-2023-gsap} or a RE dataset with wide relation coverage like SciER\cite{zhang2024scier}, there stills lacks a comprehensive fine-grained entity and relation extraction dataset.  
As a result, we see a clear need for a new comprehensive dataset of fine-grained entity types and relation types, with trustworthy manual annotation on span level to advance scholarly IE on a higher level.

\begin{table*}[t]
 \centering
 \begin{tabular}{lrrrrr}
\toprule
  & ScienceIE & SemEval18 Task 7& SciERC & SciER & \gere\\
\midrule
Annotation Unit & \ding{168}     &   \ding{171}   & \ding{171}       & \ding{170}     & \ding{170}  \\
\# Publications           & 500   & 500   & 500   & 106   & 100 \\
\# Entity Types   & 3     & -     & 6     & 3     & 10  \\
\# Relation Types & 2     & 6     & 7     & 9     & 18 \\
\# Entities       & 9,946  & 7,505   & 8,094  & 24,518 & 62,619 \\
\# Relations      & 672   & 1,583  & 4,648  & 12,083 & 35,302 \\
\# Entities in Relations & 1,824 (18\%) & 3,166 (42\%)  & 6,272  (77\%)  & 17,148 (70\%) &  46,680 (75\%) \\
\# entity Sentences & 3,319 & 2,403     & 2,551   & 7,722  & 20,020 \\ 
\# null Sentences   & 654     & 153    & 136     & 2    & 6185 \\
\# Relations/Pub.  & 3.1  & 3.3   & 9.3   & 114.0 & 353.0 \\
\# Sentences/Pub.  &   8.1  &   5.1    &   5.4    & 72.9 & 262.0  \\
\bottomrule
\end{tabular}
 \caption{
    Comparison of \gere~and 4 datasets supporting NER and RE in scientific text. Note that our dataset is the biggest dataset in terms of annotated entities and relation so far. Annotation units:
    \ding{168}=Paragraph, \ding{171}=Abstract, \ding{170}=Full Text
 }
 \label{tab:datasets_comparison}
\end{table*}

In this paper, we present \gere, a fine-grained manually annotated dataset covering 10 entity types and 18 semantically categorized relation types, focused on entities of relevance in ML-related research.
Our dataset contains mentions of 63K entities and 35K relations from all 20K sentences from the full text of 100 publications in ML and related fields. In addition, we introduce fine-tuned baseline models that demonstrate the utility of the data.

Our contributions include:
\begin{itemize}
    \item A reusable data model\footnote{We provide the data model visualization on https://data.gesis.org/gsap/gsap-ere} that captures the complex relations among ML models, datasets and tasks. We introduce a set of 18 relation labels, which are systematically organized into seven distinct semantic relation groups.
    \item We provide a dataset of more than 35K fine-grain relation annotations in 100 scientific publications, encompassing connections among ML models, methods, datasets, and tasks. Our annotations further include additional links to references and URLs, and cover all sentences in the full text of the selected publications, including those without relations or entities defined by our data model (null cases).
    \item To demonstrate the utility of our dataset, we present fine-tuned baseline models that showcase the applicability of automatic relation extraction using both pipeline and joint modeling approaches. We also report zero-shot and few-shot performance results for unsupervised prompting approaches with large language models~(LLMs).
\end{itemize}

The dataset, code for replicating the baseline models, and the annotation guideline are provided in the supplemental material and will be publicly available along with the camera-ready version. 

\section{Related Work}
\paragraph{Scholarly IE Datasets}
Multiple datasets~\cite{augenstein2017semeval,gabor2018semeval7,Luan2018SciERC,schindler2021somesci,zhang2024scier}, have been proposed as ground-truth datasets for scholarly information extraction tasks, including Named Entity Recognition (NER) and Relation Extraction (RE). We compare our \gere~with four well-established, manually annotated ground-truth datasets for scholarly IE in Table~\ref{tab:datasets_comparison}.

ScienceIE at SemEval 2017~\cite{augenstein2017semeval} and SemEval 2018 Task 7~\cite{gabor2018semeval7} are two pioneering ground-truth datasets for scholarly IE that contain both entity and relation annotations. ScienceIE includes 500 paragraphs from open access journals and contains three entity types: \textit{Task}, \textit{Method} and \textit{Material} and two relation types: \textit{Hyponym-Of} and \textit{Synonym-Of}. SemEval 2018 Task 7 provides six types of relations on annotation of entity span without typing based on abstracts from NLP publications. Another dataset SciERC~\cite{Luan2018SciERC} has been proposed, extending ScienceIE and SemEval 2018, where 500 publication abstracts from 12 AI conference or workshop proceedings are annotated. SciERC contains six entity types, including one type \textit{ Generic} supporting informal mentions of entities, and seven relation types. However, all these datasets are only based on abstracts~\cite{gabor2018semeval7,Luan2018SciERC} or selected paragraphs~\cite{augenstein2017semeval} of publications, rather than full texts, which can cover more diverse linguistic styles~\cite{zhang2024scier} and thus potentially render a wider range of entities and relations.

Therefore, full-text-based ground-truth datasets are introduced. DMDD~\cite{huitong2023dmdd} is a dataset automatically annotated with distant supervision on the full text of $\num{31219}$ scientific articles. Yet DMDD does not contain relation annotations, hence is not suitable for the relation extraction task. SciREX~\cite{jain2020scirex} contains four entity types but its relation annotations are limited to clusters of mentions rather than individual entity mention pairs, overlooking the contextual information of each entity mention. SoMeSci~\cite{schindler2021somesci} contains the full text of 100 articles from PubMed Central Open Access subset; however it focuses on various software mentions rather than ML-related entity types. SciER~\cite{zhang2024scier}, deriving from the NER dataset SciDMT~\cite{Pan2024SciDMTAL}, provides a corpus of 106 full-text annotated publications with three entity types (\textit{Dataset}, \textit{Task}, \textit{Method}) and ten relation types. To our best knowledge, SciER has the most comprehensive relation annotation scheme among the existing comparable datasets. However, it only includes explicitly named mentions, excluding the unnamed or informal mentions of entities, which accounts for a substantial amount of mentions in research publications, and results in an incomplete set of relation annotations.

Compared to these datasets, our dataset, derived from GSAP-NER~\cite{otto-etal-2023-gsap}, is based on a manual annotation of 100 full-text research publications in the fields of ML and applied ML. It contains 10 fine-grained entity types and 18 relation types. The entity label set includes both named and informal entity types, in contrast to SciER. To enable a wide range of downstream use cases, the 18 relation types cover 7 different semantic groups relating to the creation, usage, property, application and referencing of scholarly entities. Furthermore, the co-existence of flat, nested and overlapping entity mentions renders it an even more challenging relation extraction task.

\paragraph{Entity and Relation Extraction} 
Entity and relation extraction, aiming at extracting relations from text, is composed of two sub-tasks: NER and RE tasks. To achieve the end-to-end relation extraction, there exists two major paradigms:  the pipeline-based approaches~\cite{zhong2021frustratingly,ye2022plmarker} and the joined extraction approaches~\cite{luan-etal-2019-general,yan2023hgere}. 
PURE~\cite{zhong2021frustratingly} is an pipeline-based approach that uses two pre-trained encoders for the NER and RE models to capture the contextual information for each span or span pair. It also proposes an approximate batched relation model that shares the position embedding between the entity marker and the entity start and end tokens to accelerate the inference.  
PL-Marker~\cite{ye2022plmarker} is another pipeline-based approach, extending PURE batched relation model with packed levitated marker to model entity pair interaction. 
HGERE~\cite{yan2023hgere} is a state-of-the-art joint extraction approach back-boned by the PL-Marker framework \cite{ye2022plmarker} and incorporates a hypergraph neural network to facilitate information flow among entities and relations.
We fine-tune and evaluate both state-of-the-art pipeline and joint ERE models to compare their performance on \gere.

\begin{table*}[t]
 \centering
 \fontsize{9pt}{9pt}\selectfont 
\begin{tabular}{L{2cm} L{4.5cm} L{9.5cm}}
\toprule
\textbf{Semantic Group} & \textbf{Purpose} & \textbf{Example}\\
\midrule
Model Design   &   Captures internal structure and dependencies of methods and ML models. 
&  \newline  \tikzmark{start7}\colorbox{orange}{BART} follows a standard \tikzmark{end6}\colorbox{lightgray}{sequence-to-sequence Transformer} architecture.   
\begin{tikzpicture}[remember picture, overlay]
  \coordinate (C) at ($(pic cs:start7)+(4ex,2.2ex)$);
  \coordinate (D) at ($(pic cs:end6)+(16ex,2.2ex)$);
  \draw[-{Straight Barb[left]}] (C) -- (D) node[midway, above]{\small\textit{architecture}};  
 \end{tikzpicture} 
\\
\midrule
Task Binding   &  Links models and datasets to the associated tasks.   &
\newline For \tikzmark{start5}\colorbox{pink}{NER}, we consider \tikzmark{end5}\colorbox{yellow}{CoNLL-2002} (Sang, 2002) annd CoNLL-2003 (Tjong Kim Sang and De Meulder, 2003) datasets \dots
\begin{tikzpicture}[remember picture, overlay]
  \coordinate (C) at ($(pic cs:start5)+(3ex,2.0ex)$);
  \coordinate (D) at ($(pic cs:end5)+(6ex,2.0ex)$);
  \draw[{Straight Barb[right]}-] (C) -- (D) node[midway, above]{\small\textit{appliedTo}};  
 \end{tikzpicture} 
\\
\midrule
Data Usage   &  Describes how datasets are used in methods and models. & \newline To that end, we evaluate \tikzmark{start4}\colorbox{orange}{XLM-R} on the \tikzmark{end4}\colorbox{yellow}{GLUE} benchmark.  
\begin{tikzpicture}[remember picture, overlay]
  \coordinate (C) at ($(pic cs:start4)+(4ex,2.3ex)$);
  \coordinate (D) at ($(pic cs:end4)+(3ex,2.3ex)$);
  \draw[-{Straight Barb[left]}] (C) -- (D) node[midway, above]{\small\textit{evaluatedOn}};  
 \end{tikzpicture} 
\\
\midrule
Data Provenance   & Tracks the origin and transformation of datasets. & \newline \tikzmark{start3}\colorbox{olive}{PubMed} alone has a total of \tikzmark{end3}\colorbox{lime}{29M articles} as of January 2019.
\begin{tikzpicture}[remember picture, overlay]
  \coordinate (C) at ($(pic cs:start3)+(4ex,2.2ex)$);
  \coordinate (D) at ($(pic cs:end3)+(6ex,2.2ex)$);
  \draw[{Straight Barb[right]}-] (C) -- (D) node[midway, above]{\small\textit{sourcedFrom}};  
 \end{tikzpicture}  \\
\midrule
Data Properties   &  Encodes metadata of datasets (size and instance type).  &  \newline \tikzmark{start2}\colorbox{yellow}{GOOAQ} contains \tikzmark{end2}\colorbox{lime}{3 million questions} with short, snippet, or collection answers, such as the ones shown in Fig. 1.   
\begin{tikzpicture}[remember picture, overlay]
  \coordinate (C) at ($(pic cs:start2)+(4ex,2.3ex)$);
  \coordinate (D) at ($(pic cs:end2)+(10ex,2.3ex)$);
  \draw[-{Straight Barb[left]}] (C) -- (D) node[midway, above]{\small\textit{size}};  
 \end{tikzpicture}
\\
\midrule
Peer Relations   &   Covers semantic, logical and structural relation among same entity type.   &  \newline  Comparing to \tikzmark{start1}\colorbox{yellow}{WIKISQL-WEAK}, \tikzmark{end1} \colorbox{yellow}{WIKITABLE-QUESTIONS} requires more complicated reasoning capabilities. 
\begin{tikzpicture}[remember picture, overlay]
  \coordinate (C) at ($(pic cs:start1)+(4ex,2.0ex)$);
  \coordinate (D) at ($(pic cs:end1)+(15ex,2.0ex)$);
  \draw[decorate, decoration =brace] (C) -- (D) node[midway, above]{\small\textit{isComparedTo}};  
 \end{tikzpicture}
\\
\midrule
Referencing   &  Refers to external sources or documentation (citation and URL).  &  \newline The latter is inspired by \tikzmark{start0}\colorbox{orange}{TAPAS} \tikzmark{end0} (\colorbox{red}{Herzig et al., 2020}) ...

\begin{tikzpicture}[remember picture, overlay]
  \coordinate (A) at ($(pic cs:start0)+(4ex,2.4ex)$);
  \coordinate (B) at ($(pic cs:end0)+(8ex,2.4ex)$);
  \draw[-{Straight Barb[left]}] (A) -- (B) node[midway, above]{\small\textit{citation}};     
\end{tikzpicture}\\
\bottomrule

\end{tabular}

 \caption{
Brief definition of each relation semantic group and a corresponding relation example from our \gere~dataset. Note that each example only highlights entity mentions (\colorbox{orange}{MLModel}, \colorbox{yellow}{Dataset}, \colorbox{lime}{DatasetGeneric}, \colorbox{lightgray}{ModelArchitecture}, \colorbox{pink}{Task}, \colorbox{olive}{DataSource}, \colorbox{red}{ReferenceLink}) and the relation label being emphasized; other potential entities and relations may still be present in our ground truth datasets.
}
 \label{tab: datasets_semantic_groups_demo}
\end{table*}

\section{Preliminaries}
\label{sec:preliminaries}
We formalize the entity and relation extraction (ERE) task at the text span level, enabling the tracing of extracted entities and relations back to the specific sentences from which they originate.
Formally, we define the two sub-tasks, NER and RE, as follows.
We consider a collection of publications $D$ as input.
Each publication is divided into paragraphs $P = \{p_1, p_2, \dots, p_n\}$, with each paragraph further segmented into consecutive sentences $S = \{s_1, s_2, \dots, s_m\}$.
For each tokenized sentence $s_i = \{w_1, w_2, \dots, w_n\}$, we define two extraction steps to identify entities and relations as follows.
\paragraph{NER} We define the task of NER with a closed set of entity labels  $L^{\mathrm{NER}}$ as the identification of contiguous word sequences $e_i = \{w_l, \dots, w_r\}$ (i.e., spans) with a corresponding entity label from $L^{\mathrm{NER}}$.
\paragraph{RE}
In the second step, we define a closed RE task using a label set $L^{\mathrm{RE}}$, which includes a $\mathrm{NIL}$ label to denote the absence of a relation.
For every possible pair of entities $E^p = \{(e_i, e_j)| i \neq j\}$ we define the task of RE as the assignment of all pairs to one of the labels~$L^{\mathrm{RE}}$.
\paragraph{ERE} is the joined task of NER and RE from end-to-end independent of the design of the used approaches.
\paragraph{Metrics}
A joint evaluation, consistent with the task definition of ERE, encompasses both entity and relation checking.
For NER task, exact match (\textbf{\textsl{NER}})
and partial match (\textbf{\textsl{NER$\approx$}}) are two common evaluation settings. For RE, we note that its difficulty is substantially higher than that of NER. This is due to the fact that any relation pair includes two entity mentions with a relation label, where both entity mentions and relation label can lead to misalignment. 
Therefore, to support the assessment of relation annotations, we introduce four evaluation settings as below.
\begin{itemize}
    \item[-] \textbf{\textsl{RE+}} refers to both the correct labels for the entities and the relation, and exact span matching for the entities;
    \item[-] \textbf{\textsl{RE}} asks only correct relation labels and exact entity spans, with no restriction for the entity types;
    \item[-] \textbf{\textsl{RE+$\approx$}} requires exact match of all labels but allows partially overlap of entity spans;
    \item[-] \textbf{\textsl{RE$\approx$}} only asks the exact relation label and overlapped entity spans, with no further requirement.
\end{itemize}
We report micro F1 metrics, unless otherwise specified.

\section{\gere~Dataset}

In this section, we present the detailed curation process of our \gere~dataset. We reuse the comprehensive entity label sets of  GSAP-NER \cite{otto-etal-2023-gsap} and their entity annotation as a start point to create relation annotations.

\subsection{Data Collection and Processing}
As we extend and build upon the GSAP-NER \cite{otto-etal-2023-gsap} dataset with relation annotation, we first give a brief overview of GSAP-NER dataset. The GSAP-NER dataset is a corpus of 100 manually annotated full-text research publications in ML or applied ML fields. The publications are first sampled using a popularity-diversity combined strategy from HuggingFace\footnote{The used HuggingFace data was obtained from \url{https://huggingface.co/models?sort=downloads} on Nov 13, 2022.} and arXiv\footnote{The used version 123 of the arXiv dataset is available at
\url{https://www.kaggle.com/datasets/Cornell-University/arxiv/versions/123}} and then converted from PDF to plain text format via GROBID~\cite{GROBID}. The 10 entity types in this corpus are categorized into three categories: (1) ML model related, i.e. \textit{MLModel}, \textit{ModelArchitecture}, \textit{MLModelGeneric}, \textit{Method} and \textit{Task}; (2) dataset related, i.e. \textit{Dataset}, \textit{DatasetGeneric} and \textit{DataSource}; (3) miscellaneous, including \textit{ReferenceLink} and \textit{URL}.

\subsection{Data Annotation}

\subsubsection{Annotation Scheme}
We aim at a fine-grained relation label set that can reflect the comprehensive interactions among ML models, datasets, tasks, methods, model architectures and the miscellaneous entities. The interaction information should serve to improve research reproducibility and reusability, covering the provenance, usage, transformation, application and comparison of the scholarly entities. Unlike SciER \cite{zhang2024scier}, we keep the informal mentions of named entities, corresponding to entity types \textit{MLModelGeneric} and \textit{DatasetGeneric}, in the annotation scheme following SciERC \cite{Luan2018SciERC}. We argue that the informal mentions not only capture valuable information for coreferential and compositional relations, but also increase the entity-per-sentence density to cover a larger amount of relation interactions and alleviate data sparsity for certain relation types. Therefore, we define a label set of 18 relation types for seven semantic groups that are defined as follows.

\textbf{Model Design} Relations in this group capture interactions among ML models, methods and model architectures. We distinguish ML models from methods: models require executable instances, while methods refer to conceptual frameworks or techniques.
Therefore, we define \textit{usedFor} to capture a method involved in another ML model or method creation; \textit{architecture} to represent the architectural structure or component in ML model; and \textit{isBasedOn} to describe derivatives of models, e.g. a fine-tuned model based on a specific foundational model.

\textbf{Task Binding} Relation in this group captures model, method or data related entities being used for a Task entity. \textit{appliedTo} corresponds to a ML model or a method designated for some task; and \textit{benchmarkFor} describes the benchmarking between datasets and tasks.

\textbf{Data Usage} Two of the key information in ML-related publications are the training and evaluation between models/methods and datasets. Therefore, in this group, \textit{trainedOn} and \textit{evaluatedOn} capture the training and evaluation from models/methods to datasets.

\textbf{Data Provenance} Understanding data provenance for Dataset entities is crucial for reproducibility and understanding data quality and biases according to FAIR principles\footnote{https://www.go-fair.org/fair-principles/}. \textit{sourcedFrom} represents a dataset coming from some \textit{DataSource}; \textit{transformedFrom} captures a dataset transformed from another existing dataset; and \textit{generatedBy} describes a dataset generated by some method or ML model.

\textbf{Data Properties} Attributive information is also often provided when mentioning a dataset in the publication and is annotated as \textit{DatasetGeneric} according to GSAP-NER. We use \textit{size} to capture the volume or the number of entries inside datasets; and \textit{hasInstanceType} to refer the modality or more refined instance type information.

\textbf{Peer Relations } Entity mentions of the same entity type or compatible types may be co-referenced (\textit{coreference}), composed of (\textit{isPartOf}), specialized (\textit{isHyponymOf}) or compared to each other (\textit{isComparedTo}). The compatible entity types are either between the name entity type and the corresponding generic type.

\textbf{Referencing } Finally, entity mentions can be sourced from another publication or URL depending on the context. We use \textit{citation} to capture the relation between the cited entity and corresponding \textit{ReferenceLink}; and \textit{url} the relation between the referred entity and corresponding \textit{URL}.

We provide examples for each semantic group in Table~\ref{tab: datasets_semantic_groups_demo} and examples for each relation in the Appendix A.1. More detailed definitions of relations are provided in the guideline in the supplementary material.

\subsubsection{Annotation Strategy}

To ensure the annotation quality, we employ a two-phase annotation-refinement strategy. Similar to SciER~\cite{zhang2024scier} and GSAP-NER~\cite{otto-etal-2023-gsap}, we use INCEpTION~\cite{tubiblio106270} platform.

During the annotation phase, we have two student annotators with computer science background, who have finished the annotation training. Among the 100 publications, we randomly selected 10 publications for double annotation by the two annotators separately; and the remaining 90 publications are split and assigned to a single annotator each. In this phase, the annotators annotate the relations on the assigned publications with existing entity annotations. In addition, the annotators are asked to check the entity annotations from GSAP-NER~\cite{otto-etal-2023-gsap} first to ensure the relation annotations are based on the correct entity annotations.

In the refinement phase, we have two PhD students and two postdoc researchers in computer science to inspect the annotation alignment of the two annotators. Misalignment cases are filtered out in the 10 commonly annotated documents by both annotators, and patterns of misalignment on each relation are extracted. Relations in jointly annotated documents that show such misalignment patterns are re-checked and corrected if necessary. Examples of misalignment patterns are provided in the Appendix A.1.

\begin{table}[t]
    \centering
     \fontsize{9pt}{9pt}\selectfont
    \begin{tabularx}{0.84\linewidth}{lcccr}
\toprule
\textbf{Semantic Group} & \textbf{\textsl{RE+}} & \textbf{\textsl{RE}} & \textbf{\textsl{RE+$\approx$}} & \textbf{\textsl{RE$\approx$}} \\
\midrule
Task Binding &40.6 &45.6 & 44.1 & 50.0\\
Referencing & 61.6&68.1 & 65.2 & 73.7\\
Peer Relation & 57.1 & 61.4& 59.3 & 65.2\\
Model Design & 38.4 & 40.0 & 45.9 & 49.3\\
Data Usage & 61.3& 62.8 & 64.8 & 67.5 \\
Data Provenance & 51.1 & 51.5 & 58.6 & 60.0\\
Data Properties & 80.1& 80.1 & 84.6 & 84.6\\
\midrule
weighted all & 53.7 & 58.1 & 56.9 & 62.8 \\
\bottomrule
\end{tabularx}
    \caption{Interrater agreement across different relation groups measured by micro-F1 scores (\%) under different settings.}
    \label{tab:relation_IRR}
\end{table}

\subsubsection{Human Agreement and Quality Enhancement}
\label{sec: irr-def}

We calculate interrater agreement on the 10 jointly annotated publications.
We note that relation annotation is highly influenced by the underlying entity annotation. A revision of entity annotations preceding the relation annotation improves the NER interrater agreement from reported 0.61~\cite{otto-etal-2023-gsap} to 0.82 macro-F1 score under \textsl{NER} setting and 0.69 to 0.86 under \textsl{NER$\approx$} setting~\footnote{We report the full NER interrater agreement in Appendix A.1.}.
Furthermore, due to the increasing difficulty of relation annotation, we evaluate the relation annotation in four settings as described in Section \ref{sec:preliminaries}. We show the interrater agreement per semantic group in Table~\ref{tab:relation_IRR}. Despite achieving a good overall alignment, the various agreement levels in different semantic groups confirm the challenge of our relation annotation, which may be partially due to the poor reporting practices mentioned in Section~\ref{sec:intro}.  
Finally, the dataset will be publicly available along with the camera-ready version.

\section{Experiments}
We use our dataset testing models abilities to extract relational knowledge from text in the scholarly domain according to our problem definition in Section \ref{sec:preliminaries}.
\begin{table*}[t]
\centering
\begin{tabular}{llcccccc}
\toprule
\textbf{ERE-method} & \textbf{Model} & \textbf{\textsl{NER}} & \textbf{\textsl{NER$\approx$}} & \textbf{\textsl{RE}} & \textbf{\textsl{RE$\approx$}} & \textbf{\textsl{RE+}} & \textbf{\textsl{RE+$\approx$}} \\
\midrule
Supervised pipeline & PL-Marker$_{\mathrm{\gere}}$ & 72.6 & 77.7 & 41.4 & 46.2 & 36.3 & 39.9 \\
Supervised joined loss & HGERE$_{\mathrm{\gere}}$ & \textbf{80.6} & \textbf{85.8} & \textbf{54.0} & \textbf{59.8} & \textbf{46.9} & \textbf{51.3}\\
\midrule
& Qwen 2.5 32b & 42.0 & 56.9 & 7.2 & 14.6 & 7.2 & 10.9 \\
Unsupervised (LLM) pipeline & Qwen 2.5 72b & \textbf{44.4} & \textbf{59.1} & \textbf{10.1} & \textbf{15.7} & \textbf{8.2} & \textbf{11.9} \\
& LLaMA 3.1 72b & 40.5 & 55.0 & 6.4 & 9.6 & 5.7 & 7.8 \\

\bottomrule
\end{tabular}
\caption{Comparison of test set F1~score (micro average) for our baseline approaches.
Label-strict metrics are marked with~“\texttt{+}”; span-partial metrics with “$\approx$”.
LLM-based results for NER are obtained with $k=10$ examples and RE with $k=1$ example.
For both tasks we used the best performing configurations using the \emph{similar+diverse} example selection strategy.}
    \label{tab:results_approach_comparison}
\end{table*}

\subsection{Baselines}

\subsubsection{Supervised PLM-based Baselines}
Supervised entity and relation extraction (ERE) using pretrained language models (PLMs) is demonstrating state-of-the-art performance in the domain of scholarly documents~\cite{zhang2024scier}.
In our experiments we compare a pipeline-based approach, i.e. a separate NER step followed by a RE classification of the extracted entities, to a joint approach which learns to predict and classify entity mention spans and relations between pairs of detected entity mentions simultaneously using a joint loss.
\begin{itemize}
    \item \textbf{PL-Marker}~\cite{ye2022plmarker}. This pipeline-based approach first performs a span classification (NER) task, and then use the proposed Packed Levitated Marker to model and evaluate the interaction between entity mention pairs for RE.
    \item \textbf{HGERE}~\cite{yan2023hgere}. This state-of-the-art joint approach extends the PLMarker framework and incorporates a hypergraph neural network to facilitate a high-order span classification for NER and entity pair classification for RE in one step.
\end{itemize}
We fine-tune PL-Marker and HGERE on our training and validation sets and then report the performance of each fine-tuned models, denoted as \textbf{PL-Marker$_{\mathrm{\gere}}$} and \textbf{HGERE$_{\mathrm{\gere}}$}, on our test set. More specifically, we introduce and optimize a loss weight scheme in the HGERE approach in combination with \textit{ternary} information flow configuration for the hyper graph neural net (HGNN).

\subsubsection{LLM Prompting}
We run zero-shot and few-shot prompts on two state-of-the-art open-source generative LLMs, i.e. \textit{LLaMA 3.1}~\cite{touvron2023llamaopenefficientfoundation} and \textit{Qwen 2.5}~\cite{qwen2025qwen25technicalreport}.
Our prompting pipeline for NER and RE follows a two-stage process comparable to the supervised pipeline approach. For each input sentence, we first extract entities. These entities are then used to form subject-object candidate pairs, which are passed to a second prompt to classify their relation label. For $n$ predicted entities, we generate $n(n-1)$ entity pairs and classify each using an individual prompt. Example prompts used in the experiment are provided in the Appendix A.3.
 Each prompt is composed of five structured sections: \textit{Task Introduction},
 \textit{Label Definitions}, 
\textit{Step-by-Step Instructions}, \textit{Few-Shot Examples}
 and \textit{Main Input} to expect a sentence input. This structured prompt pattern is found to largely improve few-shot learning performance by helping align output generation with task expectations \cite{madaan2022textpatternseffectivechain}.

 To select the best performing hyperparameters, we explore two key prompt design parameters: the number of few-shot examples,
and the example selection strategy. We apply various hyperparameter settings to the validation set to pick the best performing configuration before applying it to the test set. 
For the example selection, we compare random selection (a fixed, randomly chosen set) to a dynamic retrieval-based strategy. The dynamic strategy can be one of:
\begin{itemize}
    \item[-] \emph{similar}, retrieving top-$k$ similar sentences via cosine similarity;
    \item [-] \emph{similar+diverse}, combining \emph{similar} with maximizing unique label re-ranking to allow broader examples.
\end{itemize}
For our experiments all examples were chosen from the training set.
We evaluate the \emph{random} and \emph{similar+diverse} strategies on the ERE task.
Both the NER and RE prompt templates are provided in the Appendix A.3. Each result was obtained from a single run per prompt.
\subsection{Experimental Setup}
\label{sec:experimental_setup}
For the supervised approaches, we fork the implementations of \citeauthor{ye2022plmarker}~\shortcite{ye2022plmarker}\footnote{\url{https://github.com/thunlp/PL-Marker}} and \citeauthor{yan2023hgere}~\shortcite{yan2023hgere}\footnote{\url{https://github.com/yanzhh/HGERE}}.
We document the steps required to setup and train the models in our forks.\footnote{The URLs of the forks will be published after the submission process.}
For the supervised methods, we use the scibert-scivocab-uncased model\footnote{\url{https://huggingface.co/allenai/scibert_scivocab_uncased}} as encoder \cite{beltagy2019scibert}.
All supervised models are tested and evaluated using 10\% of the annotated publications as validation and 10\% as test set leaving 80 annotated publications for training.
We selected for the \textit{ternary} information flow configuration for the hyper graph neural net in HGERE by hyperparameter optimization. We also tuned learning rate, batch size, and loss-weighting. Details of the optimization processes are provided in the Appendix A.3.
We run our best performing model with 5 random seeds and report the mean performance including the standard deviation for this model.

To guarantee reproducability for the unsupervised prompting approach we set the temparature parameter for all LLMs to zero.
To select the best performing hyperparameter for the unsupervised prompting approach, 
we used Qwen2.5 32B on the validation set.
%
For the final model comparison we used three open-source LLMs: \textit{Qwen2.5 (32B and 72B)} \cite{qwen2025qwen25technicalreport}, and \textit{LLaMA 3.1 70B} \cite{touvron2023llamaopenefficientfoundation} on the test set.
For efficient local inference we accessed the quantized versions of \textit{Qwen 2.5}\footnote{\url{https://ollama.com/library/qwen2.5:32b}, \url{https://ollama.com/library/qwen2.5:72b}} and
\textit{LLaMA 3.1 70B}\footnote{\url{https://ollama.com/library/llama3.1:70b-instruct-q4_K_M}} using the Ollama~\cite{ollama} Framework.
For few-shot sentence selection, we use the multi-qa-mpnet-base-cos-v1 model\footnote{\url{https://huggingface.co/sentence-transformers/multi-qa-mpnet-base-cos-v1}} from the Sentence-Transformers library \cite{reimers-2019-sentence-bert}
for sentence cosine similarity.\\
Experiments were conducted on a server running Ubuntu 22.04.4~LTS with 2× Intel Xeon 2.1~GHz CPUs (48 cores, 96 threads), 1.4~TB RAM, and 8 GPUs (4× RTX 2080~Ti with 11~GB and 4× A40 with 48~GB). The system also includes a multi-tier storage setup with both SSD and HDD arrays, totaling over 40~TB.
\subsection{Evaluation Metrics}
To comprehensively evaluate the performance of our baselines, we employ a set of metrics that capture different aspects and strictness levels for NER and RE introduced in Section \ref{sec:preliminaries}.
In general we report micro F1 performance to reflect overall model performance across labels~\cite{harbecke2022microf1classweightingmeasures}. In specific evaluation settings we report micro, macro, and weighted-averaged metrics to summarize performance for detailed comparability.
\begin{table*}[ht]
\centering
\small
\begin{tabular}{lllcccccc}
\toprule
\textbf{Example Selection} & \textbf{Setting} &\textbf{Metric} & \textbf{0}-shot & \textbf{1}-shot & \textbf{2}-shot & \textbf{5}-shot & \textbf{10}-shot & \textbf{20}-shot \\
\midrule
\multirow{6}{*}{\emph{random}}
& & micro-F1      & 19.1 & 24.7 & 23.1 & 29.7 & 34.1 & \textbf{34.4} \\
&\textsl{NER} & macro-F1      & 23.3 & 27.0 & 26.5 & 28.6 & \textbf{31.5} & 29.7 \\
& & weighted-F1    & 16.1 & 22.0 & 20.3 & 28.4 & 33.5 & \textbf{33.6} \\
\cmidrule{2-9}
& & micro-F1    &  33.0 & 41.8 & 37.1 & 50.1 & \textbf{53.3} & 50.3 \\
& \textsl{NER$\approx$} & macro-F1     & 34.5 & 40.3 & 37.7 & 43.7 & \textbf{46.0} & 40.9 \\
& & weighted-F1  & 29.8 & 39.6 & 34.5 & 49.4 & \textbf{53.1} & 49.4 \\
\midrule
\multirow{6}{*}{\emph{similar}+\emph{diverse}}
& & micro-F1      & 19.1 & 34.7 & 38.2 & 40.4 & \textbf{40.9} & 27.8 \\
& \textsl{NER} & macro-F1       & 23.3 & 34.3 & 37.6 & 39.6 & \textbf{39.7} & 21.2 \\
& & weighted-F1    & 16.1 & 34.0 & 37.9 & 40.6 & \textbf{41.4} & 28.0 \\
\cmidrule{2-9}
& & micro-F1    & 33.0 & 53.8 & 56.7 & 58.1 & \textbf{58.4} & 39.4 \\
& \textsl{NER$\approx$} & macro-F1    & 34.5 & 48.1 & 50.9 & \textbf{52.1} & 52.0 & 28.4 \\
& & weighted-F1 & 29.8 & 53.4 & 56.5 & 58.2 & \textbf{58.6} & 38.8 \\
\bottomrule
\end{tabular}
\caption{NER performance for different number of $k$-shot and example selection strategies. Micro, macro, and weighted F1 (\%;~partial and exact match) on the validation set are reported using the Qwen2.5-32B model.}
\label{tab:nerval_sum}
\end{table*}

\section{Experimental Results}
\subsection{Baseline Comparison}
Table~\ref{tab:results_approach_comparison} presents a comparative overview of the supervised pipeline, joined, and the unsupervised prompting approaches using LLMs. Our best supervised approach outperforming the LLM-based approaches by a large margin with an \textsl{NER} F1 performance of $80.6\pm0.3\%$ and \textsl{RE+} F1 of $46.9\pm0.5\%$. We report the best results for each setting with optimized hyperparameters. Specifically, the LLM-based results are all based on \emph{similar+diverse} strategy, with 10-example for NER evaluations and 1-example for RE evaluations. We detail the hyperparameter selection experiments in Section \ref{subsec: hyperparameter_exp}.

We first observe a substantial performance advantage of supervised models over unsupervised LLM prompting approaches, ranging $18.6$--$39.8\%$ for \textsl{NER}/\textsl{NER$\approx$} and $28.1\%$--$50\%$ for all four RE settings.
Then comparing the two supervised PLM approaches, the joint approach HGERE outperforms the pipeline approach PL-Marker under every setting for both NER and RE, demonstrating the robustness of this start-of-the-art method.
Finally, among LLM prompting results, Qwen 72b achieves the best performance for all settings. However, compared to the two supervised PLM approaches, this advantage of Qwen 72B over other LLMs is marginal, in particular for the RE task, marking more research effort is needed for LLMs on such a comprehensive fine-grained IE task.
Our findings are consistent with similar observations reported in related work~\cite{zhang2024scier}.\\
When comparing runtime performance between the supervised and unsupervised approaches, the PLM-based method outperforms the LLM-based methods by a substantial margin.
On the test corpus of 10 documents, inference with the LLM approach is 182 times slower than with the fine-tuned PLM on the same hardware described in Section~\ref{sec:experimental_setup} (4 minutes vs. 12 hours and 29 minutes), while training the PLM required 2 hours and 30 minutes for a single training run.

\subsection{Impact of LLM Few-Shot Hyperparameters}
\label{subsec: hyperparameter_exp}
For our unsupervised prompting approach we run our pipeline on the validation set under different configurations to pick the optimized number of few-shot examples ($k$) and example selection strategy.  Table~\ref{tab:nerval_sum} shows the results on NER in these configurations. We include the detailed performance in Appendix A.4 for further fine grained analysis. We observe that the consistently best performing configuration for NER is $k=10$ examples with \emph{similar+diverse} example selection, achieving 58\% F1 score, +5\% compared to random example selection. The performance sharply declines after 20 examples. For RE, we run $k\in \{0,1,2,5\}$ with \emph{similar+diverse} strategy. As shown in Table~\ref{table:re_val}, the overall RE performance is unsatisfying, but 1-shot and 2-shot achieve the best performance with minimal difference among themselves.
Taking these analysis into consideration, We select 10-shot \emph{similar+diverse} for NER and 1-shot \emph{similar+diverse} for RE as the optimized hyperparameters, based on the trade-off between performance and computational cost. 
The detailed results on the test set for pipeline-based NER and RE are provided in Appendix A.4. 

\begin{table}[t]
\centering
\small
\begin{tabular}{cccccc}
\toprule
\textbf{$k$-Shot} & \textbf{\textsl{RE$\approx$}} & \textbf{\textsl{RE+$\approx$}} & \textbf{\textsl{RE}} & \textbf{\textsl{RE+}} \\
\midrule

0 & 16.8 & 11.7 & 8.9 & 6.5 \\

1  & 20.4 & \textbf{14.4} & 10.7 & 8.0 \\
2 & \textbf{20.5} & \textbf{14.4} & \textbf{10.8} & \textbf{8.1} \\
5 & 19.9 & 14.0 & 10.3 & 7.6 \\
\bottomrule

\end{tabular}
\caption{
Micro F1 scores (\%) for the RE task on the validation set using Qwen2.5-32B.
}
\label{table:re_val}
\end{table}

\section{Conclusion}
We introduced \gere, a manually annotated dataset designed for fine-grained scholarly information extraction in ML research. It features 10 detailed entity types and 18 relation types annotated across full-text scientific articles, comprising more than 62K annotated entities and 35K relations. Both data model and dataset shed light on the interdependencies between key concepts involved in machine learning-related research and facilitate research into information extraction methods that can provide a more structured view on machine learning-related artefacts and their relations at scale. 
Our experiments show that fine-tuned models trained on this dataset significantly outperform prompting methods using LLMs, highlighting the current limitations of unsupervised approaches in domain-specific IE tasks. The significant performance gap supports our claim that curated datasets like \gere~are essential to advance reliable extraction of research metadata, enabling downstream applications such as knowledge graph construction or the understanding of reproducibility of ML-based research at scale.
\section{Limitations}
The first limitation is the field of publications in our dataset. We include only ML and applied ML fields to slightly simplify the annotation task without resorting to domain experts. We emphasize the complexity of RE annotation, and will transfer the data curation lessons when we extend the corpus domain. The second limitation is our corpus covers only sentence-level annotations. Although our sentence-level ERE dataset is already challenging due to the multiplicity of entity mentions and relations, future work for a document-level corpus will benefit more scholarly IE task on long texts.


\section*{Acknowledgments}
We thank the anonymous reviewers for their constructive feedback.
This work has been partially funded by the Deutsche Forschungsgemeinschaft (DFG, German Research Foundation) as part of the Projects BERD@NFDI (grant number 460037581), NFDI4DS (grant number 460234259), as well as Unknown Data (grant number 460676019).
\bibliography{references}

\end{document}